\documentclass[lettersize,journal]{IEEEtran}
\usepackage{amsmath,amsfonts,amssymb}
\usepackage{algorithmic}
\usepackage{algorithm}
\usepackage{array}
\usepackage[caption=false,font=footnotesize,labelfont=rm,,textfont=rm]{subfig}
\makeatletter
\def\endthebibliography{%
  \def\@noitemerr{\@latex@warning{Empty `thebibliography' environment}}%
  \endlist
}
\makeatother
\usepackage{textcomp}
\usepackage{stfloats}
\usepackage{verbatim}
\usepackage{graphicx}
\usepackage{cite}

\usepackage{mathrsfs}
\usepackage{bbm}
\usepackage{booktabs}
\usepackage{balance}
\usepackage{multirow}
\usepackage{footnote}
\usepackage{color}
\usepackage{xcolor}
\usepackage{diagbox}
\usepackage{enumitem}
\newcommand{\etal}{\textit{et al}. }
\newcommand{\ie}{\textit{i}.\textit{e}., }

\usepackage{url}
\usepackage{bm}
\usepackage{threeparttable}
\usepackage[colorlinks,linkcolor=blue,urlcolor=blue,citecolor=blue]{hyperref}

\begin{document}

\title{Multicam-SLAM: Non-overlapping Multi-camera SLAM for Indirect Visual Localization and Navigation}

\author{Shenghao~Li\textsuperscript{1},
		Luchao~Pang\textsuperscript{2},
        Xianglong~Hu\textsuperscript{3}
\thanks{\textsuperscript{1} S. Li is with the Department of Automation, School of Electronic Information and Electrical Engineering, Shanghai Jiao Tong University, Shanghai 200240, P.R China (email: lch94102@sjtu.edu.cn).}
\thanks{\textsuperscript{2} L. Pang is with School of Mechanical and Power Engineering, East China University of Science and Technology, Shanghai 200237, P.R China (e-mail: lucien\_pang@163.com).}
\thanks{\textsuperscript{3} X. Hu is an independent researcher. (email: hxianglong@gmail.com)}
}

\markboth{Journal of \LaTeX\ Class Files,~Vol.~14, No.~8, August~2021}%
{Shell \MakeLowercase{\textit{et al.}}: A Sample Article Using IEEEtran.cls for IEEE Journals}

\IEEEpubid{0000--0000/00\$00.00~\copyright~2021 IEEE}

\maketitle

\begin{abstract}
This paper presents a novel approach to visual simultaneous localization and mapping (SLAM) using multiple RGB-D cameras. The proposed method, Multicam-SLAM, significantly enhances the robustness and accuracy of SLAM systems by capturing more comprehensive spatial information from various perspectives. This method enables the accurate determination of pose relationships among multiple cameras without the need for overlapping fields of view. The proposed Muticam-SLAM includes a unique multi-camera model, a multi-keyframes structure, and several parallel SLAM threads. The multi-camera model allows for the integration of data from multiple cameras, while the multi-keyframes and parallel SLAM threads ensure efficient and accurate pose estimation and mapping. Extensive experiments in various environments demonstrate the superior accuracy and robustness of the proposed method compared to conventional single-camera SLAM systems. The results highlight the potential of the proposed Multicam-SLAM for more complex and challenging applications. Code is available at \url{https://github.com/AlterPang/Multi_ORB_SLAM}.
\end{abstract}


\section{Introduction}\label{Introduction}

Simultaneous Localization and Mapping (SLAM) is a critical technology that has found extensive applications in various fields such as autonomous navigation, augmented reality, and robotics. Among the various types of SLAM, laser SLAM has been widely used due to its high accuracy. However, it comes with high costs and lacks the ability to extract semantic information from the environment. On the other hand, visual SLAM, which uses cameras as the primary sensor, offers cost advantages and the ability to extract rich semantic information from imagery. 

Despite these advantages, visual SLAM, particularly when using a single camera system, has its limitations. The field of view is constrained and greatly influenced by ambient lighting and textural variations in the environment. This makes the system prone to tracking losses due to inadequate visual features in environments lacking distinct features, such as plain white walls, or in complex settings with rapid changes in lighting or visual texture.

To overcome these challenges, researchers have explored various strategies. One promising direction in the field of SLAM has been the integration of multiple sensors. For instance, laser-vision fusion combines the high-resolution color information from cameras with the accurate distance measurements from laser sensors~\cite{liu2023multi}. Similarly, inertial-vision fusion integrates the high-frequency motion data from inertial measurement units (IMUs) with the visual data from cameras, providing robust and accurate pose estimation even in dynamic environments~\cite{mur2017visual}. Another approach involves employing lenses with broader angles of view, such as fisheye cameras. However, this approach leads to significant image distortion and loss of features, rendering it less suitable for RGB-D cameras, which offer superior benefits in indoor settings compared to monocular cameras due to their ability to directly capture depth information.

In this paper, we propose a novel approach to enhance the robustness and accuracy of visual SLAM by deploying multiple RGB-D cameras as the multicam entity. The proposed methodology enables the arbitrary configuration of multiple RGB-D cameras oriented in various directions. This means that the cameras can be placed in any spatial arrangement and orientation, providing flexibility in capturing diverse viewpoints. Since this setting do not require overlapping fields of view between cameras, each camera can cover a unique part of the environment, thereby maximizing the spatial coverage. A novel calibration method is also proposed to estimate the relative rigid transformation between non-overlapping cameras on-the-fly. This approach significantly enriches the visual features available for more robust and accurate pose tracking and mapping.
\IEEEpubidadjcol

The main contributions of this paper are threefold:

\begin{enumerate}
    \item We propose a novel multi-camera visual SLAM method that leverages multiple RGB-D cameras to capture more comprehensive spatial information, thereby improving the system's performance in complex environments. 
    \item We introduce a calibration method based on pose graph optimization for the multicam entity. This method is designed to calibrate the relative poses of cameras configured without overlapping fields of view, addressing a key challenge in multi-camera SLAM systems.
    \item We conduct extensive experiments in various environments to validate the effectiveness and robustness of the proposed method. The results demonstrate the superior performance of our method over existing single-camera SLAM methods, highlighting its potential for more complex and demanding applications.
\end{enumerate}

\section{Related Work} \label{RelatedWork}

The integration of vision in robotics has dramatically enhanced the performance of localization and navigation tasks. Recent advancements have seen the introduction of numerous sophisticated Simultaneous Localization and Mapping (SLAM) systems that employ RGB-D cameras. These systems have proven to be efficient and accurate in obtaining camera pose and robot motion.

KinectFusion~\cite{newcombe2011kinectfusion} was a pioneering system capable of real-time 2D reconstruction using RGB-D cameras. It leverages point clouds derived from depth images for camera pose estimation via the Iterative Closest Point (ICP) algorithm. However, it does not utilize RGB data, which limits the reconstruction to scenes of predetermined size and lacks capabilities for loop closure detection. RTAB-MAP~\cite{labbe2013appearance}, a groundbreaking RGB-D SLAM methodology, incorporates a short-term, working, and long-term memory management mechanism. This mechanism reduces the required number of nodes for graph optimization and loop closure detection, ensuring real-time performance and accuracy, especially in large scenes.
ElasticFusion~\cite{whelan2015elasticfusion} utilizes both the color consistency of images and point clouds generated from depth images for ICP to estimate camera poses. However, it is limited to reconstructing small-scale environments, such as a single room.
RGBD-SLAMv2~\cite{endres20133} is known for its comprehensive capabilities in RGB-D SLAM, but it is hindered by substantial computational demands and limited real-time performance.

ORB-SLAM and its improved versions~\cite{mur2015orb, mur2017orb, ORB-SLAM3} are highly refined and sophisticated open-source SLAM systems based on features and keyframes. They support monocular, stereo, and RGB-D cameras, demonstrating excellent real-time capabilities and robust loop closure detection algorithms.
The foundational algorithm proposed in this paper is grounded in the principles of ORB-SLAM, which is noted for its real-time performance capabilities and superior loop closure detection algorithms.

Along with the rapid development and wide applications of deep learning and VLMs~\cite{liu2024infrared, jiang2021recurrent, zheng2024advanced, yu2024credit, Shen2024Harnessing}, there are several promising approaches to enhance the robustness of VSLAM, such as our previous work that adopts learned features~\cite{li2022ssfslam} to enhance the tracking stability and accuracy of feature-based VSLAM under challenging conditions. Recently, neural SLAM systems~\cite{li2024representing} have been proposed to establish implicit reconstructions by optimizing neural representation functions on-the-fly. In this paper, we approach the improvement of VSLAM robustness from a different angle by using the multicam entity, which is a complementary method to the above mentioned techniques.

There has been extensive research focusing on adapting single-camera SLAM techniques to multi-camera configurations, thereby expanding the application spectrum of visual SLAM across various domains such as autonomous vehicles and drones. This adaptation is crucial for enhancing the spatial awareness and navigational capabilities of autonomous systems in complex environments.

Heng et al.~\cite{heng2015leveraging} introduced a high-precision, structure-based method for the extrinsic calibration of multicam entitys using high-performance SLAM to create detailed maps that facilitate the calibration process. This approach demonstrates significant advancements in the accuracy and reliability of multi-camera setups in dynamic environments. Häne et al.~\cite{hane20173d} applied multi-camera visual SLAM to enhance vehicle positioning and navigation. They proposed a novel SLAM system that operates effectively with multiple fisheye cameras possessing non-overlapping fields of view. Their method involves a two-step calibration process that initially uses wheel odometers for establishing a preliminary map, which is then refined using structural data to achieve dense map reconstruction.

In addition to automotive navigation, a significant application area for multi-camera visual SLAM is in unmanned aerial vehicles (UAVs)~\cite{yang2015visual, yang2017multi, heng2015self, harmat2012parallel}. These systems are increasingly leveraging multi-camera configurations to enhance spatial awareness and operational safety during flight. Saowu et al. have made substantial contributions to advancing visual SLAM by employing two monocular cameras to robustly and autonomously locate and navigate drones~\cite{yang2014visual, scherer2014robust, yang2017multi}. These cameras are strategically mounted to face forward and downward, optimizing visibility and navigational accuracy.

Harmat et al. pioneered the adaptation of PTAM to multi-camera SLAM, implementing it on drones within indoor settings to achieve enhanced tracking robustness~\cite{harmat2012parallel}. This advancement was further refined to produce more accurate map points, which in turn improved pose estimations and map precision~\cite{harmat2015multi}. Their developments were made available to the wider research community through open sourcing their algorithms. Tribou et al.~\cite{tribou2015multi} significantly enhanced Harmat's methodology by implementing a parameterization and initialization scheme that facilitated rapid initialization and reliable tracking and mapping with only a few initial images from the camera array.

Building on their earlier work~\cite{heng2015leveraging}, Heng et al.~\cite{heng2015self} utilized a stereo camera to impart crucial scale information to other cameras within a multi-fisheye camera system. They also introduced an innovative motion estimation technique that requires merely three points to accurately determine the pose of the multi-camera setup, significantly reducing the computational complexity typically associated with such processes. MultiCOL-SLAM~\cite{urban2016multicol} further extends the monocular ORB-SLAM algorithm to accommodate multi-fisheye camera SLAM configurations at any angle, utilizing the previously proposed MultiCOL model, which exhibits robust performance in backend graph optimization. However, the predominant use of fisheye cameras introduces substantial computational overhead if image correction is undertaken. Conversely, extracting ORB features from highly distorted, uncorrected original images often results in matching failures, indicating a trade-off between computational efficiency and algorithmic robustness.

The focus of this work is on SLAM research that employs multiple RGB-D cameras, a less explored area in the field. While numerous studies aim at map reconstruction using multiple RGB-D cameras, they often do not cover the entire SLAM process, primarily focusing on achieving high mapping accuracy in constrained environments~\cite{liu2019novel,perez2017extrinsic}. Our research also addresses the crucial aspect of robotic localization and navigation, which is frequently overlooked in these studies.

Recent studies have significantly advanced research on multi RGB-D camera calibration. Shaowu~\etal~\cite{yang2015visual} proposed an iterative optimization method for pose tracking and map optimization in a multi RGB-D camera SLAM system based on the PTAM algorithm. They provided a detailed mathematical analysis of the process and developed a semi-automatic method for calibrating the extrinsic parameters of multi RGB-D cameras with non-overlapping fields of view. However, their research did not address the critical issue of loop closure detection, leaving a gap in the field. Notably, Meng~\etal~\cite{meng2018dense} proposed two calibration methods based on visual odometry and SLAM, integrating these techniques to significantly enhance calibration accuracy. They further extended the capabilities of existing technologies by adapting ElasticFusion for multicam entitys to perform dense scene reconstruction. However, in the SLAM processes they outlined, each camera's pose is independently tracked, and the most accurate pose is used as a reference to correct the others. This approach enhances tracking robustness but introduces system redundancy and limits the method's suitability to small scene 3D reconstructions due to inherent limitations of ElasticFusion. Moreover, the effectiveness of loop closure detection in the proposed algorithm was not evaluated, raising concerns about the overall utility of the enhancements.

Another pivotal study~\cite{chen2018calibrate} focused on calibrating multi RGB-D cameras using a high-precision laser scanner to capture environmental point clouds. They then employed the ICP algorithm to calibrate the extrinsic parameters of a three Kinect camera system. Although this method simplifies operations by transforming data from additional cameras into the calibrated camera's coordinate system, it does not fundamentally enhance SLAM performance. The map accuracy heavily depends on the precision of the extrinsic calibration of the camera array, which is a significant limitation.

In contrast to the methods discussed above, the multi-camera SLAM system proposed in this paper is more comprehensive and robust, covering the entire SLAM process, including pose tracking, mapping, and loop closure detection. The multicam entity is integrated into the complete SLAM process, and the multi-camera model is used in graph optimization to define the reprojection error and derive Jacobian matrices for both bundle adjustment and pose graph optimization, employing the inter-camera constraints. This paper also introduces an innovative automatic calibration method for multi-camera extrinsics with non-overlapping fields of view that can be oriented in any direction, effectively applying this system to visual SLAM. This development not only boosts the robustness and precision of SLAM pose tracking but also significantly enhances backend optimization processes, resulting in more accurate mapping outcomes. The integration of multiple cameras into the SLAM framework notably increases the efficiency of the mapping process, showcasing the potential of multicam entitys in broadening the operational scope and reliability of SLAM technologies.


\section{Methodology} \label{Methodology}
\begin{figure}[t]
	\centering
	\includegraphics[width=\columnwidth]{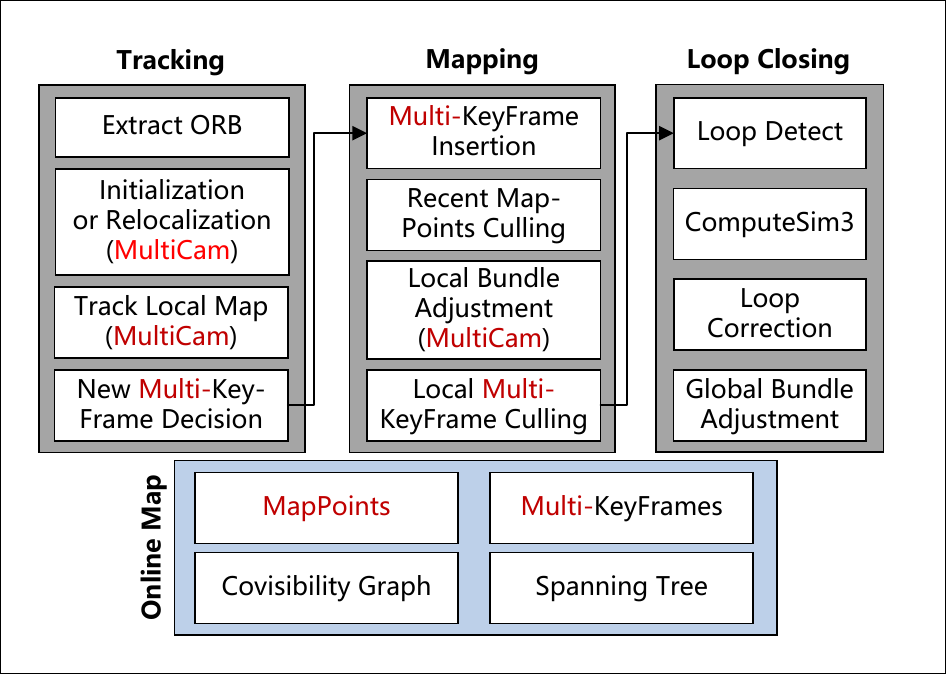}
	\caption{Overview of the Proposed Multicam-SLAM system. The system is structured into three concurrent threads: pose tracking, local mapping, and loop closure detection.}
	\label{fig1-overview}
\end{figure}

The proposed Multicam-SLAM, aliased Multicam-SLAM, is integrated as a multi-keyframe (MKF) system. The multi-keyframe contains multiple image captured by different cameras and the corresponding local features extracted, and we name the structure of multiple cameras as the multicam entity (MCE). The MKF system is structured into three concurrent threads, each handling distinct tasks: pose tracking, local mapping, and loop closure detection. These threads are illustrated in Fig.~\ref{fig1-overview}. The keyframes are defined as a set of images captured by multiple views during the exploration, incorporating 3D map points back-projected from different cameras. As the multicam entity captures the scene from various viewpoints, we introduce a comprehensive multi-camera model to methodically address the structural and mapping challenges associated with the multi-camera setup. 

\subsection{Scene Exploration}

Upon the arrival of a new frame, which consists of images from multiple cameras, the tracking thread promptly detects keypoints and computes descriptors for each image to facilitate the subsequent feature matching. The system initially attempts to predict the current poses based on the prior frames and then optimizes these predictions in real-time as the initial poses. If unsuccessful, the system advances to perform feature matching between the current frame and the latest reference multi-keyframe, securing the pose  estimation through rigorous pose graph optimization. To handle the tracking loss issue, the system employs EPnP and RANSAC to relocalize using all of the reference multi-keyframes. We adopt the ORB features~\cite{rublee2011orb} for their efficiency in feature matching, ensuring the system's real-time performance and accuracy.

Once the initial poses are determined, the system continues to monitor the local map and to project additional match points into 3D space. In the meantime, the system contemplates whether to integrate the current frame as a new multi-keyframe into the SLAM system, subsequently transferring this multi-keyframe to the mapping thread. Upon receipt of a new multi-keyframe by the mapping thread, it methodically removes recently generated map points that do not conform to a predefined quality criteria. This thread then triangulates matched 2D keypoints to cultivate new map points from the newly added adjacent multi-keyframes, as delineated by the covisibility graph. Triangulation is conducted not only between images from a single camera but also across images from different cameras, facilitating enhanced structural understanding and mapping accuracy. Subsequently, local bundle adjustment optimizes the poses of multi-keyframes and map points within the local map, ensuring high-fidelity reconstruction.

Moreover, the mapping thread assesses the redundancy of multi-keyframes and decides whether to prune them from the map to optimize system efficiency. Each newly integrated multi-keyframe triggers the loop closure detection thread to scan the map for potential loops using the bag-of-words (BoW) model. This process involves comparing the BoW scores of the current frame with historical multi-keyframes to ascertain previously visited scenes and identify loop closure candidates. Upon loop detection, the system calculates a similarity transformation to commence loop correction, followed by essential graph optimization and global bundle adjustment to enhance the system's accuracy and robustness.

In the backend, the system leverages the G2O framework for sophisticated graph optimization processes. Differing from conventional single-camera visual SLAM, this research employs a graph optimization technique tailored to multicam entitys to formulate an accurate objective function. This includes deriving Jacobian matrices essential for both bundle adjustment and pose graph optimization within the multicam entity, thus redefining the optimization framework.

\begin{figure}[t]
	\centering
	\includegraphics[width=\columnwidth]{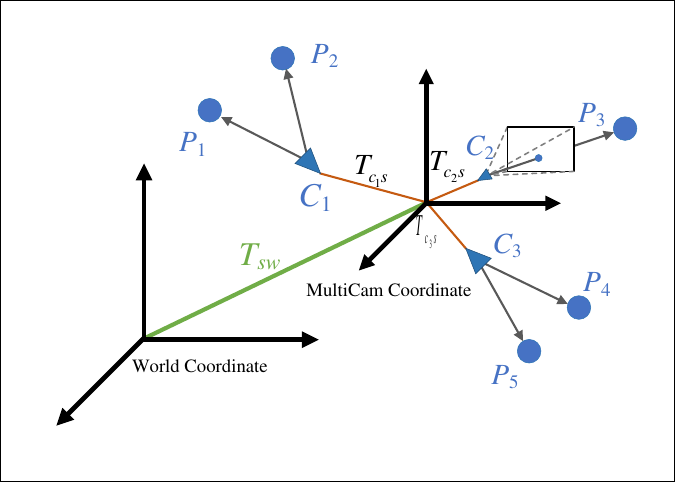}
	\caption{Illustration of the Multiple Camera Coordinate System.}
	\label{fig2-multicamera-system}
\end{figure}

\subsection{Multi-Camera Mapping}

The structure of a multicam entity is more complex compared to a single camera. To address the mapping problem in Multicam-SLAM, we introduce a multi-camera model. This model is applicable to multicam entitys with non-overlapping fields of view and configured with arbitrary cameras at any position and angle. To flexibly configure the coordinate systems of multiple cameras, we establish an multi-camera coordinate system. Points in the scene are not directly transformed from the world coordinate system to each camera coordinate system, but are first transformed into this multi-camera coordinate system. 

In the multi-camera model, the pose of camera $C_i$, denoted as $T_{c_i}$, is obtained by transforming the pose of the multicam entity $\mathbf{T}_{sw}$ as:
\begin{equation}
\boldsymbol{T}_{c_i}=\boldsymbol{T}_{c_i s} \cdot \boldsymbol{T}_{s w}
\end{equation}
where $\boldsymbol{T}_{c_i s}$ is the $4 \times 4$ transformation matrix from the multi-camera coordinate system to the coordinate system of camera $C_i$, which is the external parameter in the multicam entity. To simplify the model, we set the multi-camera coordinate system to coincide with the coordinate system of the camera, \ie $\boldsymbol{T}_{c_1 s}$ is set to an identity matrix.

Simultaneously, a map point $\mathbf{P}_j$ in the world coordinate system projected onto the image plane of a camera $C_i$ can be represented as:
\begin{equation}
\boldsymbol{u}_{i j}=\pi_i\left(\boldsymbol{T}_{c_i s} \cdot \boldsymbol{T}_{s w} \cdot \boldsymbol{P}_j\right)=\pi_i\left(\boldsymbol{P}_{c_i j}\right)
\end{equation}
where $\pi_i$ is the projection model of camera $C_i$, $\boldsymbol{T}_{c_i s}$ is the transformation matrix from the multi-camera coordinate system to the coordinate system of camera $C_i$, which includes a rotation matrix $\boldsymbol{R}_{c_i s}$ and a translation vector $\boldsymbol{t}_{c_i s}$. Meanwhile, $\boldsymbol{T}_{sw}$ represents the transformation from the world coordinate system to the multi-camera coordinate system. As shown in Fig.~\ref{fig2-multicamera-system}, consider a multicam entity composed of three cameras. The blue triangle represents a camera in the multicam entity, the red solid line represents the pose of each camera in the multicam entity, and the green solid line represents the pose transformation of the multicam entity relative to the origin of the world coordinate system.

During the exploration, 3D map points are obtained by the back-projection of matched local feature. Considering the pixel coordinate of a matched keypoint $u_{c_i}$ on camera $C_i$, its 3D world coordinate can be calculated as:
\begin{equation}
\boldsymbol{P}_w=\boldsymbol{T}_{s w}^{-1} \cdot \boldsymbol{T}_{c_i s}^{-\mathbf{1}} \cdot \pi_{c_i}^{-1}\left(\boldsymbol{u}_{c_i}, Z\right)
\end{equation}
where $\boldsymbol{P}_w$ is the 3D map point in the world coordinate system, $\pi_{c_i}^{-1}$ is the back-projection function of $C_i$, and $Z$ is the depth of $u_{c_i}$.

\subsubsection{Map Representation}
The map in this paper primarily consists of 3D map points, multi-keyframes (MKFs), and a covisibility Graph.

\paragraph{Map Points}
3D map points are the most basic entities in SLAM. They are obtained either directly by back-projection of ORB features based on depth information provided by RGB-D cameras in SLAM, or by triangulation calculations between multiple keyframes. Each map point $\mathbf{P}_i$ contains the following components:
\begin{enumerate}
    \item 3D coordinates of the map point, represented in world coordinates as $\mathbf{P}_i = [X_i, Y_i, Z_i]^T$.
    
    \item ORB descriptor with the lowest Hamming distance to all other matched 2D local features.
    
    \item The maximum distance $d_{\text{max}}$ and minimum distance $d_{\text{min}}$ at which the map point can be observed. These distances are used to reduce map point amount in local map tracking.
    
    \item Observation angle $\boldsymbol{n}_i = [n_x, n_y, n_z]^T$, which is the unit vector of the average direction of all camera views that can observe this map point. The observation angle is adopted to determine whether to move the map point.
\end{enumerate}

\paragraph{Multi-Keyframes}
As the main frame structure of the proposed Multicam-SLAM, the multi-keyframe includes the following components:

\begin{enumerate}
\item The pose of the multi-keyframe $\boldsymbol{T}_{sw}$, which is the pose of the multicam entity (MCE), used for transforming map points between the world coordinate system and the multi-camera coordinate system.

\item The intrinsic and extrinsic parameters of each camera within the multicam entity, $\boldsymbol{T}_{cis}$. These two parameters are used for the projection and back-projection between 3D map points and 2D keypoints, as well as for transforming between cameras.

\item Local features extracted from all the cameras of this multi-keyframe. Each local feature contains its keypoint coordinates and descriptor on the corresponding camera.

\item BoW vector used for scene recognition of multi-keyframes. This vector is calculated by a pretrained bag-of-word model based on local features (ORB), and the similarity between two multi-keyframes is calculated by comparing the cosine similarity of their BoW vectors.
\end{enumerate}

\paragraph{Covisibility Graph}

The covisibility graph is used to represent the observations of shared 3D map points among multi-keyframes. Derived from~\cite{endres20133}, the covisibility graph is conceptualized as a weighted undirected acyclic graph, wherein each node embodies a multi-keyframe and each edge represents the covisibility between multi-keyframes. The edge weight, denoted as $\mathbf{W}$, quantifies covisible map points between two multi-keyframes. We establish an edge if the number of covisible map points exceeds a threshold $W_{\text{min}}=20$. 

The connected multi-keyframes in the covisibility graph are referred to as covisible multi-keyframes, and the top 20 frames with the most covisible points are referred to as the neighboring multi-keyframes. Due to the use of a non-overlapping multicam entity, the covisibility relationship that exists in one camera among multi-keyframes may not exist in other cameras. This can lead to isolated multi-keyframes, increasing the difficulty of loop closure detection and reducing the robustness of the system. Therefore, we modify the strategy for selecting neighboring frames. After selecting a certain number of frames (set to 20 in this paper) with the highest degree of covisibility as the neighboring multi-keyframes of the current frame, if a camera of the current frame has a covisibility relationship with other multi-keyframes, but these multi-keyframes are not selected as neighboring frames, then the frame with the most common points among these multi-keyframes is selected as an additional neighboring keyframe.

At the start of visual SLAM, map initialization is required to obtain the initial pose and to establish the map. During the initialization of monocular ORB-SLAM, the fundamental matrix $\mathbf{F}$ and homography matrix $\mathbf{H}$ are calculated simultaneously between two frames, and the suitable model for initialization is selected by comparing the scores of $\mathbf{F}$ and $\mathbf{H}$. When the camera only performs rotational motion or only observes a planar scene, the homography matrix $\mathbf{H}$ is prefered. Otherwise, the fundamental matrix $\mathbf{F}$ is more reliable, and the scene is initialized by reconstructing the camera pose and scene points. For RGB-D cameras, the map is initialized by back-projecting the local feature coordinates and depth information of the first frame, eliminating the extensive calculations required for monocular camera initialization and allowing more flexible acquisition of the initial map. After initialization, this frame is added as the initial multi-keyframe to the local mapping thread, and the system officially enters the pose tracking process. Moreover, since we use a multi-RGB-D camera system with non-overlapping fields of view, the map needs to be initialized separately in each camera and then fused through the given multi-camera model to obtain the initial map.

\subsection{Multi-Camera Calibration} \label{subsec-multicam-calibration}

Given the essential requirement for precise pose relationships between cameras in visual SLAM systems with multiple cameras, it becomes imperative to calibrate the external matrices $\mathbf{T}_{c_i s}$ as discussed in the previous section. To address this, we propose a calibration method that uses pose graph optimization to calibrate the multicam entity without an additional calibration process. A simplified SLAM procedure is first run, during which the multi-RGB-D camera system is rotated to ensure a certain amount of overlap data between different cameras. Then, ORB feature matching is performed across different cameras, and the ICP algorithm is adopted to calculate the transformation matrix $\mathbf{T}$ between the matched frames. Finally, the relative poses between the cameras in the entity is refined through pose graph optimization.

\begin{figure}[t]
	\centering
    \subfloat[(a)]{
		\includegraphics[width=0.46\columnwidth]{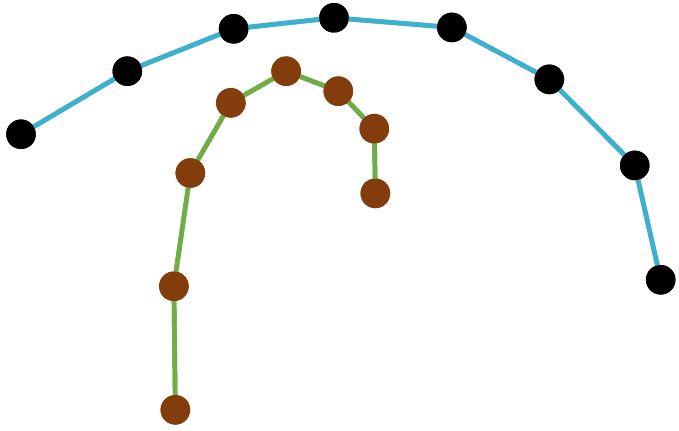} \label{fig3-pose-graph-a}
	}
	\subfloat[(b)]{
		\includegraphics[width=0.46\columnwidth]{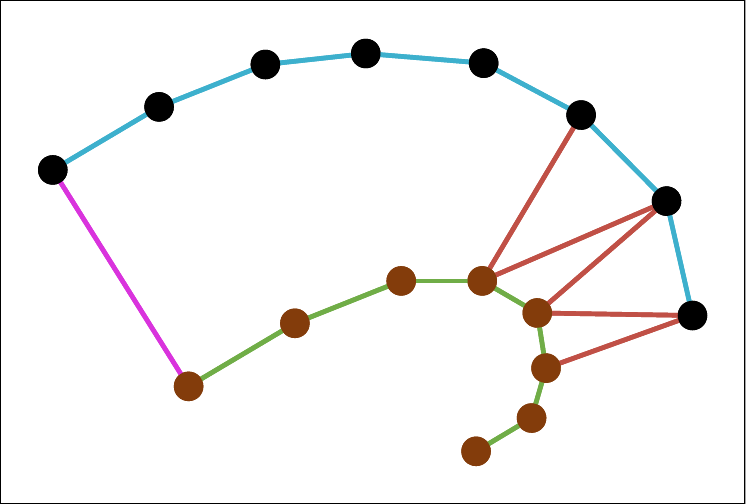} \label{fig3-pose-graph-b}
	}
	\caption{Pose Graph Optimization for multicam entity. (a) The initial pose graph. The black and brown vertices represent the poses of the two cameras respectively. The blue and green edges are the transformations between adjacent key frames of the two cameras respectively. (b) Matching-based Edge Establishment. The red edges represent The matching keyframes, the purple edge represents the final extrinsic parameters between the initial keyframes of the two cameras.}
	\label{fig3-pose-graph}
\end{figure}

More specifically, in this simplified SLAM procedure, each camera independently performs tracking and mapping, respectively. The keyframe pose of each camera is taken as a vertex, and the transformation between keyframes is taken as an edge to construct the pose graph. In this way, each camera obtains its initial pose graph. Taking two cameras as an example, as shown in Fig.~\ref{fig3-pose-graph}, the black and brown vertices in the figure represent the keyframe poses of the two cameras, and the blue and green edges are the transformations between adjacent keyframes of the two cameras. Then, all keyframes of one camera are traversed and feature matched with each keyframe of the other camera. If the number of local feature matches is greater than 10, it is considered a successful match, and an edge is added between these two vertices in the pose graph. The constraint of the edge is the pose transformation between the two frames obtained by using the ICP algorithm. The objective function of the ICP algorithm can be written as:
\begin{equation}
\boldsymbol{e}=\frac{1}{2} \sum_{i=1}^N\left\|\boldsymbol{p}_i-\boldsymbol{T} \boldsymbol{p}_i^{\prime}\right\|^2
\end{equation}
where $\boldsymbol{p}_i$ and $\boldsymbol{p}_i^{\prime}$ are the matched feature pairs, and $N$ is the total number of matching points. 

Finally, the camera poses are optimized with constraints represented as the pose graph. In the multicam entity, one camera is selected as the main camera, and the world coordinate system is set according to the starting pose of the main camera. Edges are established across all the camera of the multicam entity, and the corresponding rigid transformation estimated by ICP is used as the initial variable for optimization. Then, with the starting pose of the main camera fixed, the constructed pose graph is optimized with respect to the relative poses between the cameras of the multicam entity. Therefore, the camera poses relative to the main camera are obtained. As shown in Fig.~\ref{fig3-pose-graph-b}, the red edges represent the matched keyframes, and the purple edges represent the external parameters between the initial keyframes of the two cameras obtained in the end.

\subsection{Graph Optimization of multicam entity}

The graph optimization of the multicam entity includes both pose graph optimization and Bundle Adjustment (BA) for multiple cameras. BA optimizes the poses of multiple cameras along with map points, while the pose optimization focuses solely on the poses within the multicam entity.

Upon the insertion of a new multi-keyframe, this multi-keyframe and its neighboring multi-keyframes that share a covisibility relationship become local multi-keyframes. The poses of these local multi-keyframes and the poses of their associated map points are subjected to local BA optimization. In addition, when the system detects and corrects a loop closure, a global BA optimization thread is initiated in the backend to optimize all multi-keyframes and map points on the map.

This paper proposes a multi-camera BA optimization algorithm for optimizing the poses of the multicam entity and map points. In order to perform the optimization, the Jacobian matrix of the error function is required. The error function is defined as the re-projection error of the map points in the image plane of the camera, and the Jacobian matrix can be calculated by the chain rule, which is defined as:
\begin{equation}
\frac{\partial \mathbf{e}}{\partial \boldsymbol{\delta} \boldsymbol{\xi}}=\frac{\partial \mathbf{e}}{\partial \mathbf{P}_m} \frac{\partial \mathbf{P}_m}{\partial \boldsymbol{\delta} \boldsymbol{\xi}}=\frac{\partial \mathbf{e}}{\partial \mathbf{P}_{c_i}} \frac{\partial \mathbf{P}_{c_i}}{\partial \mathbf{P}_m} \frac{\partial \mathbf{P}_m}{\partial \boldsymbol{\delta} \boldsymbol{\xi}},
\end{equation}
where $\boldsymbol{\delta}$ and $\boldsymbol{\xi}$ are keypoint 2D coordinates and camera poses to be optimized, respectively, $\mathbf{P}_m$ is the map point, and $\mathbf{P}_{c_i}$ is the map point projected onto the image plane of camera $C_i$. The Jacobian matrix of the error function is calculated by the chain rule, and the final Jacobian matrix is obtained by combining the Jacobian matrices of the re-projection errors.

RGB-D cameras are treated as virtual stereo cameras to be compatible int the optimization. The error $\mathbf{e}$ in BA optimization is the pixel coordinate error of the map points and the horizontal coordinate error on the virtual right camera. Therefore, for the camera $C_i$ in the multicam entity, we have:
\begin{equation}
\begin{aligned}
& u=f_x \frac{X_{c_i}}{Z_{c_i}}+c_x, v=f_y \frac{Y_{c_i}}{Z_{c_i}}+c_y \\
&u_r=f_y \frac{X_{c_i}-b}{Z_{c_i}}+c_x.
\end{aligned}
\end{equation}
Then we have:
\begin{equation}
\frac{\partial \mathbf{e}}{\partial \mathbf{P}_{c_i}}=\left[\begin{array}{ccc}
-\frac{f_x}{z_c} & 0 & \frac{f_x x_c}{z_c^2} \\
0 & -\frac{f_y}{z_c} & \frac{f_y y_c}{z_c^2} \\
-\frac{f_x}{z_c} & 0 & \frac{f_x x_c-b_f}{z_c^2}
\end{array}\right]
\end{equation}

For the second term, we have \( \mathbf{P}_{c_i} = \mathbf{R}_{c_im} \mathbf{P}_m + \mathbf{t}_{c_im} \). The third term is the derivative of the point transformed to the multi-camera coordinate system with respect to the Lie algebra, which is:
\begin{equation}
\begin{aligned}
\frac{\partial \mathbf{P}_m}{\partial \boldsymbol{\delta} \boldsymbol{\xi}}&=\left[\begin{array}{cc}
-\mathbf{P}_m \wedge & I \\
0^{\mathrm{T}} & 0^{\mathrm{T}}
\end{array}\right] \\
&=\left[\begin{array}{cccccc}
0 & z_m & -y_m & \multirow{3}{*}{\huge{I}} \\
-z_m & 0 & x_m &  \\
y_m & -x_m & 0 & 
\end{array}\right].
\end{aligned}
\end{equation}

Therefore, the Jacobian matrix can be written as:
\begin{equation}
\begin{aligned}
\mathbf{J}_{1}=\frac{\partial \mathbf{e}}{\partial \boldsymbol{\delta} \boldsymbol{\xi}} = \frac{\partial \mathbf{e}}{\partial \mathbf{P}_{c_i}} \mathbf{R}_{c_im} \frac{\partial \mathbf{P}_m}{\partial \boldsymbol{\delta} \boldsymbol{\xi}}.
\end{aligned}
\end{equation}

In addition, it is necessary to optimize the position of the map points, so we need to know the derivative of $e$ with respect to the map point $\mathbf{P}_w$. This is the second Jacobian matrix:
\begin{equation}
\mathbf{J}_{2}=\frac{\partial \mathbf{e}}{\partial \mathbf{P}_w}=\frac{\partial \mathbf{e}}{\partial \mathbf{P}_{c_i}} \frac{\partial \mathbf{P}_{c_i}}{\partial \mathbf{P}_m} \frac{\partial \mathbf{P}_m}{\partial \mathbf{P}_w},
\end{equation}
where the first and second terms on the right side of the equation have been derived earlier. The third term can be written as:
\begin{equation}
\mathbf{P}_m=\exp \left(\boldsymbol{\xi}^{\wedge}\right) \mathbf{P}_w=\mathbf{R}_{m w} \mathbf{P}_w+\mathbf{t}_{m w}.
\end{equation}

\section{Experiments} \label{Experiments}

To evaluate Multicam-SLAM proposed in this paper, we designed versatile experiments to compare it against a single-camera SLAM. The experiments consist of two parts:

\begin{enumerate}
\item Mobile platform experiment: the multicam entity is installed on a mobile robot to test the performance of Multicam-SLAM during planar movement. To verify the robustness of Multicam-SLAM in scenes with sparse local features, a tracking robustness experiment was conducted in addition.

\item Handheld experiment: experimental personnel handheld the multicam entity to test the effect of Multicam-SLAM during 3D movement. This experiment also recorded the time consumed by each thread and compared it with the single-camera method.
\end{enumerate}

The experiments evaluated the robustness and accuracy of the visual SLAM by calculating the tracking rate and the discrepancy between the camera trajectories and the ground truth. In the trajectory accuracy experiment, each sequence was run 10 times, and the median was taken.

\subsection{Experimental Setup}
\subsubsection{Hardware Setting}
\begin{figure}[t]
	\centering
	\includegraphics[width=0.77\columnwidth]{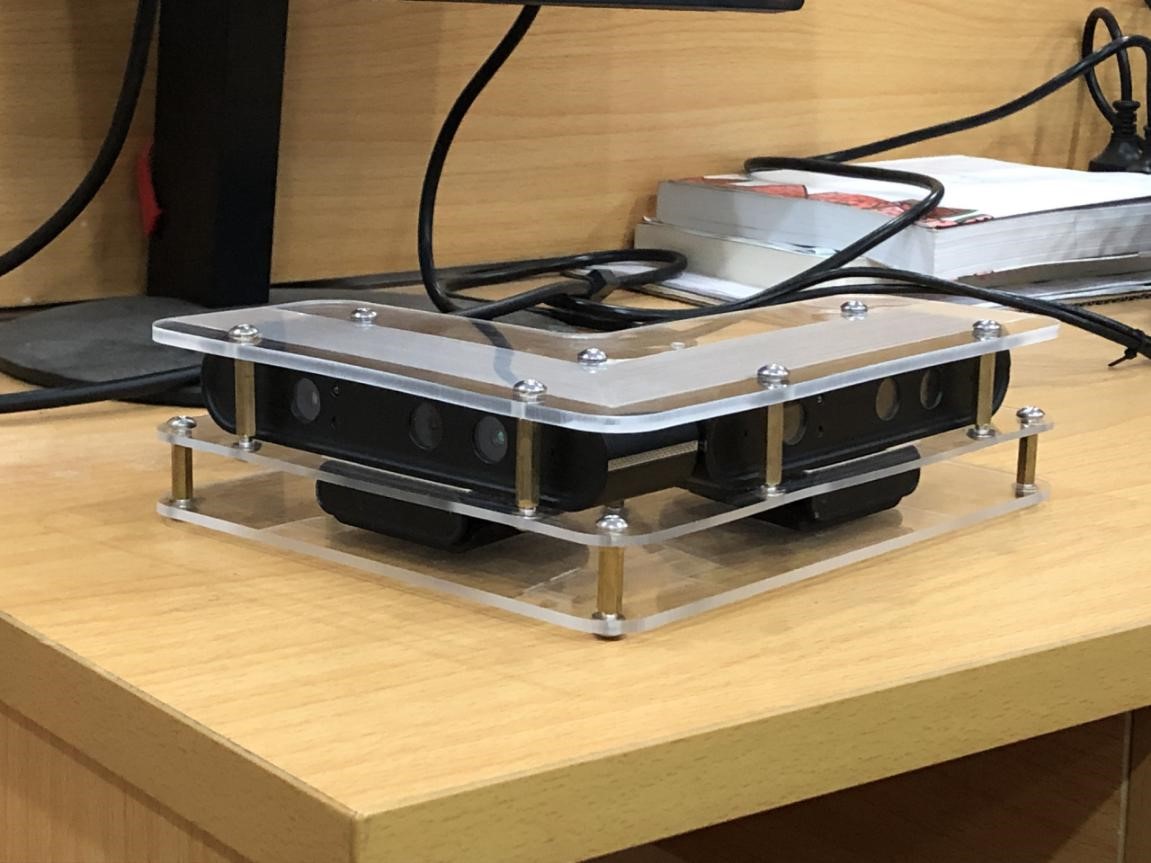}
	\caption{Illustration of the Multiple Camera Hardware.}
	\label{fig4-exp-multicamera-hardware}
\end{figure}

The experiments used Orbbec Astra RGB-D cameras providing $640 \times 480$ images at 30 FPS. We used two RGB-D cameras structured as a rigid with one facing forward and the other facing right, as shown in Fig.~\ref{fig4-exp-multicamera-hardware}. The relative poses between cameras are calibrated as described in Section~\ref{subsec-multicam-calibration}.

\begin{figure}[t]
	\centering
	\includegraphics[width=\columnwidth]{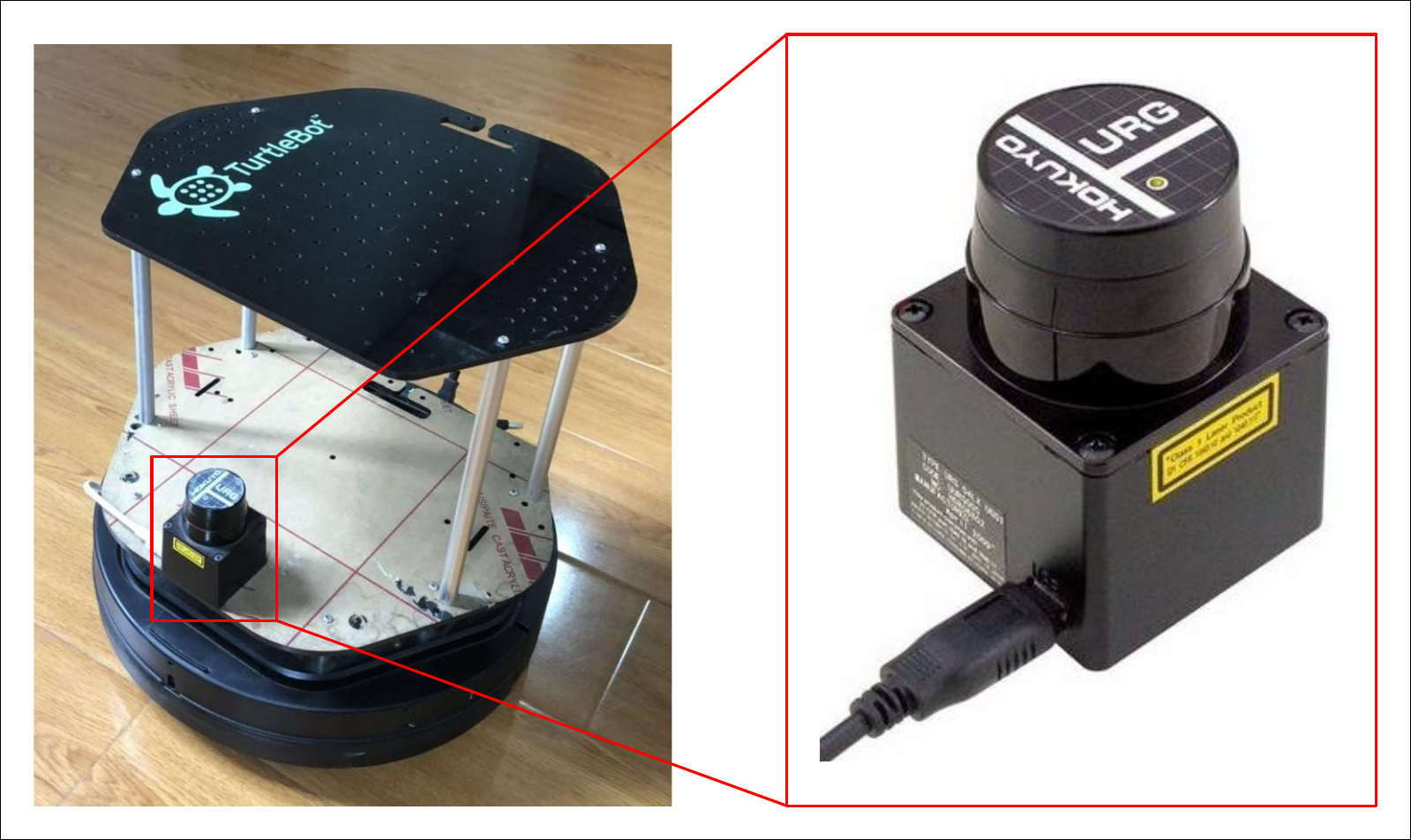}
	\caption{Illustration of the Mobile Platform for Evaluation.}
	\label{fig5-exp-mobile-platform}
\end{figure}

In the mobile platform experiment, a Turtlebot robot was adopted as the main platform, as shown in Fig.~\ref{fig5-exp-mobile-platform}. The ground truth trajectory is obtained by a hardware odometry refined by a North Star laser radar. In the handheld experiment, the VICON optical motion capture system was used to capture the pose of the multicam entity in real time as the true trajectory data. The experiment is performed on a laptop with Intel Core i5-7300HQ with 8G memory, and the open-source Robot Operating System (ROS) is used to keep the experimental settings aligned throughout the evaluation.

\subsubsection{Baseline Method}
To provide a fair evaluation regarding the multi-camera setting, we adopt ORB-SLAM2 as the baseline method because of its similar parallel structure to the proposed Multicam-SLAM with a single camera. ORB-SLAM2~\cite{mur2017visual} is one of the most complete feature-based and keyframe-based visual SLAM system. It supports monocular cameras, stereo cameras, and RGB-D cameras, and has excellent loop closure detection and backend optimization modules. All comparative experiments in this paper will be conducted with RGB-D cameras, while ORB-SLAM2 only receive the frames from the main camera.

In order to reduce experimental misalignment, the experiments are conducted by collecting sequences and running visual SLAM offline. The multicam entity is used to collect videos in multiple scenes, and external equipment such as laser radars and motion capture systems are adopted to provide the ground truth trajectories. Due to hardware limitations, the frame rate of the collected sequence is lower than the camera frame rate, approximately 20 frames per second.

\subsubsection{Evaluation Metrics}

To evaluate the accuracy of the SLAM system, the experiment compares the difference between the estimated camera pose $T_t$ and the real pose $T_t^{gt}$ at timestep $t$. Since the frequency of pose data collection is inconsistent with the frequency of image data collection, there is usually a slight time difference between $T_t$ and $T_t^{gt}$, which requires the use of timestamps to associate the estimated pose with the real pose. The difference between pose $T_t$ and pose $T_t^{gt}$ at time $t$ can be represented by the transformation matrix:
\begin{equation}
\boldsymbol{T}_t^{rel}=\boldsymbol{T}_t^{g t^{-1}} \boldsymbol{T}_t
\end{equation}

We uses the root mean square error (RMSE) of the absolute trajectory error (ATE) to evaluate the trajectory error. Before calculating the absolute error, the trajectories need to be aligned using the similarity transformation $S$:
\begin{equation}
\boldsymbol{T}_t^{r e l}=\boldsymbol{T}_t^{g t^{-1}} \boldsymbol{S} \boldsymbol{T}_t
\end{equation}
For $N$ pairs of poses, the root mean square error of the absolute trajectory error is calculated as follows:
\begin{equation}
\mathrm{ATE}=\sqrt{\frac{1}{N} \sum_{t=1}^N\left\|\operatorname{trans}\left(\boldsymbol{T}_t^{rel}\right)\right\|^2}
\end{equation}
where $\operatorname{trans}\left(\boldsymbol{T}_t^{rel}\right)$ represents the translation part of the transformation matrix $\boldsymbol{T}_t^{rel}$.

In addition to the trajectory accuracy, the tracking stability is another essential metric to evaluate the robustness of VSLAM systems. With a higher stability, the system is less likely to be affected by external perturbation. We adopt the tracking rate to represent the tracking stability. The tracking rate is the proportion of frames successfully tracked to all the frames during the visual SLAM exploration. The calculation formula is:
\begin{equation}
\text { Tracking Rate }=\frac{N_{\text {tracked }}}{N_{\text {sum }}}
\end{equation}
where $N_\text{tracked}$ represents the frame number successfully tracked by the system, and $N_\text{sum}$ represents the total frame amount.

\subsection{Mobile Platform Experiment}


To verify the accuracy and robustness of the proposed Multicam-SLAM in planar motion, we conduct mobile platform experiments in various indoor scenes. During the experiments, the multicam entity was installed on a mobile robot platform. In the experiment, the experimenter manually manipulated the mobile robot to move in multiple experimental scenes and collected multi RGB-D camera sequences.

Different sequences were collected multiple times in different experimental environments, and each sequence was run 10 times using both ORB-SLAM2 and the proposed Multicam-SLAM. During the exploration, the image data of all cameras were used as the input for Multicam-SLAM, while only the data of the forward-facing main camera was inputed for ORB-SLAM2.


The mobile platform experiments were conducted on all six sequences using the two methods, and Table~\ref{table-01-robot-exp-ate} records the average root mean square error of the trajectories after each method was run 10 times on each sequence.

As reported in Table~\ref{table-01-robot-exp-ate}, the first four sequences were collected in an indoor environment without dynamic object interference. When collecting the sequences, the robot's motion in the room1 and room2 sequences formed a loop. Therefore, the trajectory errors on these sequences are relatively smaller. Regarding the ATEs and tracking rates of room1 to room4, Multicam-SLAM shows a similar stable tracking rate to ORB-SLAM2 in small and ideal indoor scenes. In the meantime, Multicam-SLAM has a higher trajectory accuracy on all four sequences.

\begin{figure}[t]
	\centering
	\includegraphics[width=\columnwidth]{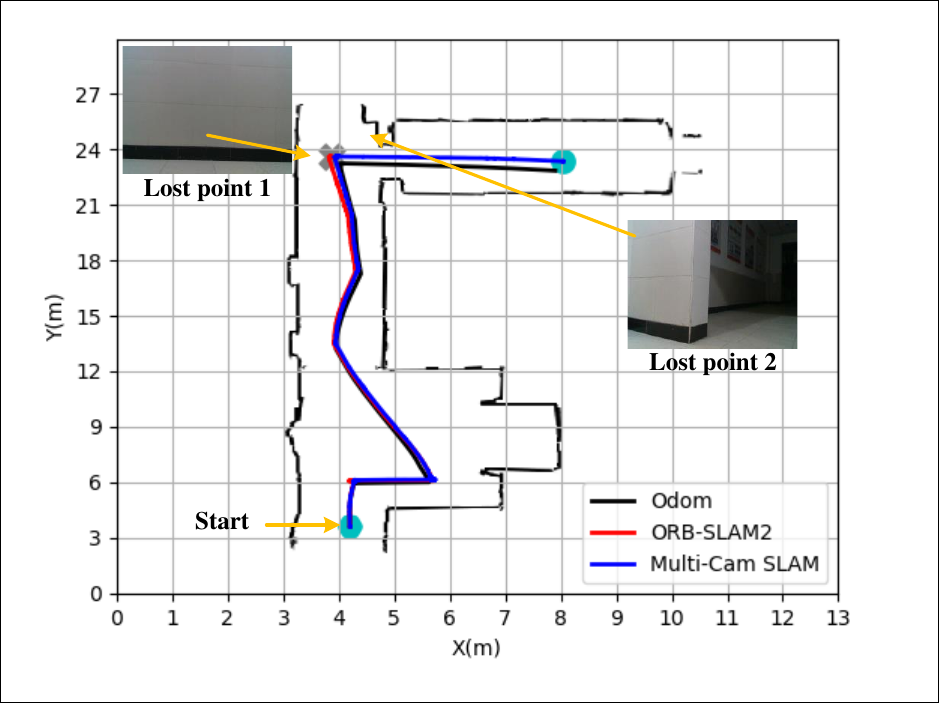}
	\caption{Visualization of Camera Paths in the Mobile Platform Experiment on the Corridor1 Sequence.}
	\label{fig7-exp-corridor-path}
\end{figure}

\begin{table}[t]
	\caption{Evaluation Results on Mobile Platform}
	\label{table-01-robot-exp-ate}
	\begin{center}
		\begin{threeparttable}
			\setlength{\tabcolsep}{1.5mm}
			\begin{tabular}{l c c c c}
				\toprule
				\multirow{2}{*}{Sequence}			& \multicolumn{2}{c}{ATE [mm] $\downarrow$} & \multicolumn{2}{c}{Tracking Rate $\uparrow$}\\
                \cmidrule(lr){2-3} \cmidrule(lr){4-5}
				&ORB.S2~\cite{mur2017orb} &MultiCam.S &ORB.S2~\cite{mur2017orb} &MultiCam.S\\
				\midrule
                room1        &62.8	    &{\color{blue}\textbf{56.0}}	    &99.90\%	&100.00\% \\
                room2        &72.5	    &{\color{blue}\textbf{54.2}}	    &100.00\%	&100.00\% \\
                room3        &146.3	    &{\color{blue}\textbf{100.1}}	    &88.54\%	&88.54\% \\
                room4        &164.4	    &{\color{blue}\textbf{112.4}}	    &99.58\%	&99.52\% \\
                corridor1        &{\color{blue}\textbf{303.7}}	    &360.4	    &84.90\%	&99.87\% \\
                corridor2        &—	            &{\color{blue}\textbf{331.8}}	    &21.07\%	&{\color{blue}\textbf{94.04\%}} \\
				\bottomrule
			\end{tabular}
			\begin{tablenotes}
				\item[*] The better methods are marked as {\color{blue}\textbf{blue}} (close rates not marked).
			\end{tablenotes}
		\end{threeparttable}
	\end{center}
\end{table}

The sequences corridor1 and corridor2 were collected in a "20x2" meter narrow corridor, as shown in Fig.~\ref{fig7-exp-corridor-path}. In the corridor1 sequence, the trajectory error of ORB-SLAM2 is lower than that of Multicam-SLAM. This is because ORB-SLAM2 lost tracking midway, leaving only the early high-accuracy trajectory, as shown in Fig.~\ref{fig5-exp-mobile-platform}. The tracking loss location is where the robot turns right from the bright part of the corridor to the dark part. The reason for the tracking loss is that when the camera turns from the bright part of the corridor to the dark part, the exposure changes, the visual features are unstable, and the SLAM system cannot track the local features, resulting in tracking failure. In the corridor2 sequence, the robot's trajectory is opposite to that of corridor1. ORB-SLAM2 lost tracking at the beginning of the run when the robot first turned, with a tracking rate of only 21.07\%. The loss location is where the robot faces the corridor wall and turns left, as shown in Fig.~\ref{fig7-exp-corridor-path}. 

There are two reasons for this phenomenon. First, the wall surface is smooth and has less texture, so the system cannot extract enough local features, making tracking easy to lose. Second, the field of view of the RGB-D camera is narrow. Even if the system can extract certain local features on the wall, the observed local features are easy to move outside the camera view during rotation if it is close to the wall, causing the system to be unable to effectively track the local features. However, Multicam-SLAM was able to maintain a high tracking rate in both of the aforementioned sequences, due to the increased feature observations and more reliable pose optimization.

\begin{table}[t]
	\caption{Tracking Rate Against Limited Texture}
	\label{table-02-wall-tracking}
	\begin{center}
		\begin{threeparttable}
			\setlength{\tabcolsep}{2.5mm}
			\begin{tabular}{l c c c c}
				\toprule
				\multirow{2}{*}{Method}	& \multirow{2}{*}{Dist.W}	& \multicolumn{3}{c}{TR @ Rot $\uparrow$} \\
				\cmidrule(lr){3-5}
				&&0.8 rad/s &0.5 rad/s &0.25 rad/s\\
				\midrule
                ORB-SLAM2        &0.5 m	&0.0\%	&45.0\%	&100.00\% \\
                Multicam-SLAM    &0.5 m	&15.0\%	&85.0\%	&100.00\% \\
                ORB-SLAM2        &1.0 m	&0.0\%	&45.0\%	&100.00\% \\
                Multicam-SLAM    &1.0 m	&10.0\%	&90.0\%	&100.00\% \\
				\bottomrule
			\end{tabular}
			\begin{tablenotes}
				\item[*] The better methods are marked as {\color{blue}\textbf{blue}}. Dist.W denotes the distance between the camera and the wall. TR @ Rot denotes the tracking rate at different rotation speeds.
			\end{tablenotes}
		\end{threeparttable}
	\end{center}
\end{table}

To further validate the robustness of the proposed method in close proximity to wall surfaces, a wall tracking experiment was conducted. After the initialization, the mobile robot was made to rotate facing a white wall, and 20 runs were conducted at different distances from the wall and at different rotation speeds. In each experiment, if the robot rotated 90° without losing tracking, it was considered successfully tracked. Table~\ref{table-02-wall-tracking} summarizes the probability of successful tracking under various conditions.

The results show that when the robot's rotation speed is low, both ORB-SLAM2 and Multicam-SLAM proposed in this paper can fully track the camera motion. Because the wall surface is not completely smooth and textureless, and visual SLAM can still track a small number of local features. When the robot's rotation speed increased to 0.5 rad/s, there was a significant difference in the tracking rate between these two methods. ORB-SLAM2 could still maintain a high tracking rate of 85\% at a distance of 1.0 meter from the wall, but the rate dropped significantly at 0.5m, with only 9 successful runs. In contrast, Multicam-SLAM was still able to maintain a high tracking rate of 90.0\% and 80.0\% respectively. 

When the robot's rotation speed continued to increase to 0.8 rad/s, the tracking rate of both methods dropped significantly, with ORB-SLAM2 even having 0 successful instances at 0.5 meters. The above experimental comparison effect is due to the fact that the multi-camera can capture more local features, so it can track better during the rotation process, proving the robustness improvement brought by the multi-camera's expanded field of view. Note that when the rotation speed is relatively fast, the tracking rate of Multicam-SLAM also decreases, which is due to the excessive computation and the inability to keep up with the speed, this problem can be resolved by further parallel computing.

Through the analysis of the results of the SLAM experiments and tracking experiments on the above six sequences, it can be found that under ideal or challenging conditions, Multicam-SLAM algorithm proposed in this paper has a higher accuracy and an improved robustness compared to the single-camera method.

\subsection{Handheld Camera Experiment}
\begin{figure}[t]
	\centering
	\includegraphics[width=\columnwidth]{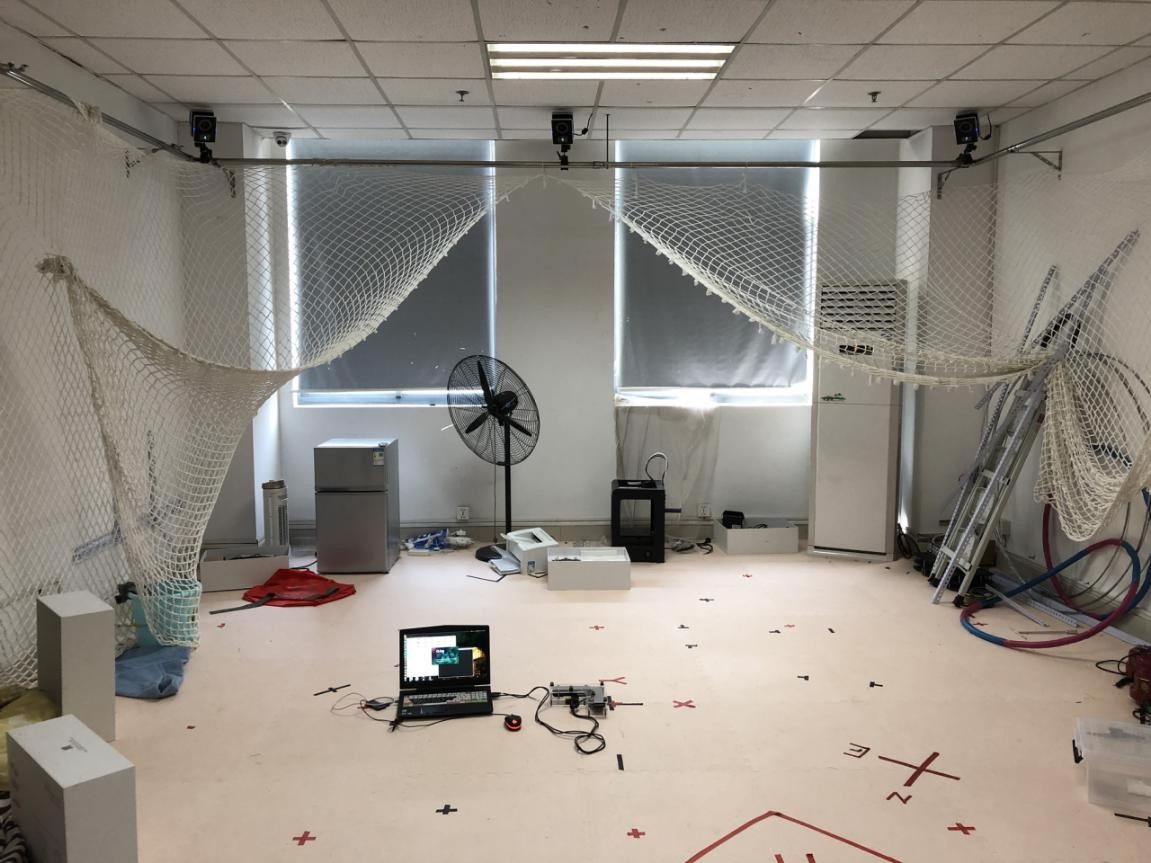}
	\caption{Illustration of the Motion Capture System and the Experimental Scene of the Handheld Evaluation.}
	\label{fig8-exp-motion-capture}
\end{figure}

\begin{table}[t]
	\caption{Evaluation Results on Handheld Setting}
	\label{table-03-handheld-exp-ate}
	\begin{center}
		\begin{threeparttable}
			\setlength{\tabcolsep}{1.2mm}
			\begin{tabular}{l c c c c}
				\toprule
				\multirow{2}{*}{Sequence}			& \multicolumn{2}{c}{ATE [mm] $\downarrow$} & \multicolumn{2}{c}{Tracking Rate $\uparrow$}\\
				\cmidrule(lr){2-3}\cmidrule(lr){4-5}
				&ORB.S2~\cite{mur2017orb} &MultiCam.S &ORB.S2~\cite{mur2017orb} &MultiCam.S\\
				\midrule
                lab\_hand1	&56.8	&{\color{blue}\textbf{56.0}}	&100.00\%	&99.47\% \\
                lab\_hand2	&73.8	&{\color{blue}\textbf{54.3}}	&99.72\%	&100.00\% \\
                lab\_hand3	&{\color{blue}\textbf{76.7}}	&79.6	&89.21\%	&{\color{blue}\textbf{100.00\%}} \\
                lab\_hand4   &72.9	&{\color{blue}\textbf{62.1}}	&90.35\%	&{\color{blue}\textbf{100.00\%}} \\
				\bottomrule
			\end{tabular}
			\begin{tablenotes}
				\item[*] The better methods are marked as {\color{blue}\textbf{blue}} (close rates not marked).
			\end{tablenotes}
		\end{threeparttable}
	\end{center}
\end{table}

In this experiment, we manually operated the multicam entity in indoor scenes, while using the high-precision VICON optical motion capture system to obtain the ground-truth trajectory of the multicam entity in real time. Unlike the mobile platform experiment, the multicam entity in this experiment will move in a 3D space with six degrees of freedom, and the camera poses and motion speed may change dramatically during the exploration, placing high demands on tracking robustness. A total of 4 different trajectory sequences were collected.

The experimental environment is shown in Fig.~\ref{fig8-exp-motion-capture}. The laboratory is equipped with nets on all sides with repetitive visual features with similar patterns, which further introduces challenges for the accuracy and robustness of pose tracking and loop closure detection. On the other hand, the net-like geometric features make it easy to interfere with the depth measurement of RGB-D cameras. The absolute trajectory error and tracking rate resutls are reported in Table~\ref{table-03-handheld-exp-ate}.

\begin{figure*}[t]
	\centering
	\subfloat[Lab\_hand3]{
		\includegraphics[width=0.9\columnwidth]{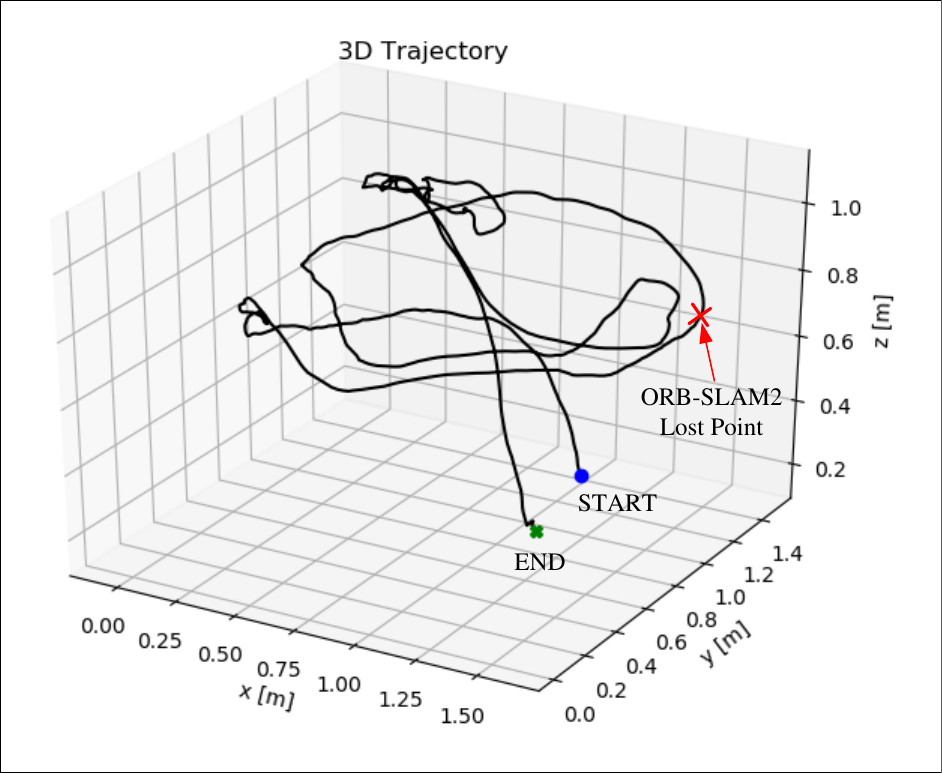} \label{fig9-exp-3d1}
	}
	\subfloat[Lab\_hand4]{
		\includegraphics[width=0.9\columnwidth]{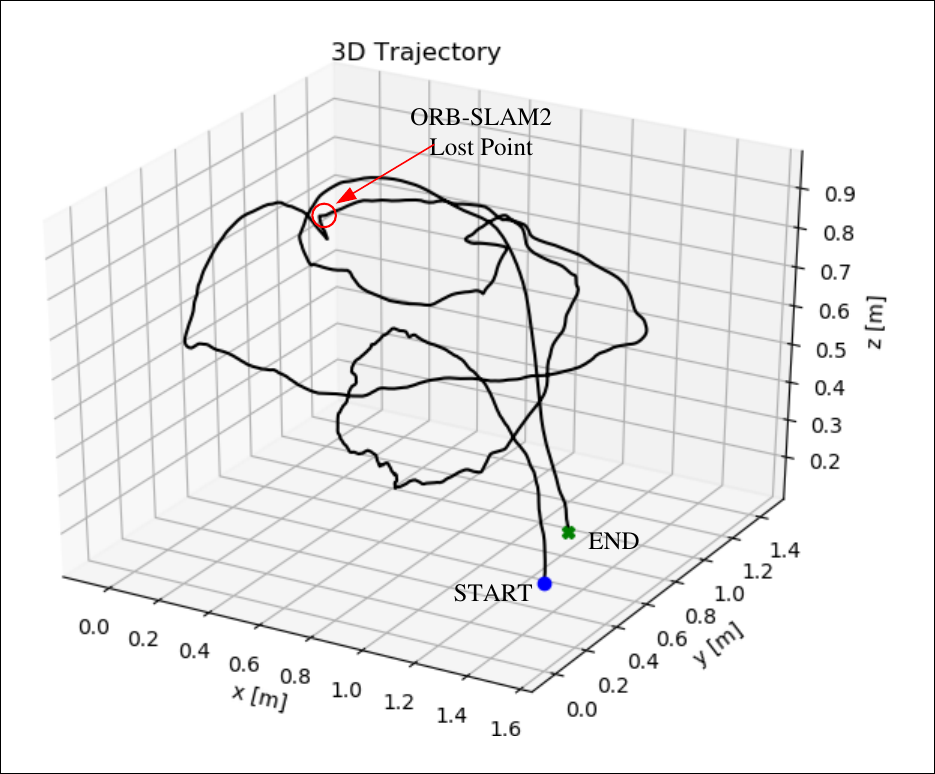} \label{fig9-exp-3d2}
	}
	\caption{Visualization of the Camera Paths in the Handheld Experiment.}
	\label{fig9-exp-3d}
\end{figure*}

In terms of tracking robustness, Multicam-SLAM shows high tracking rates on all sequences. Except for the slightly lower tracking rate on the lab\_hand1 sequence than ORB-SLAM2, the tracking rate of Multicam-SLAM on other sequences all reached 100\%. However, ORB-SLAM2 lost tracking on the lab\_hand3 and lab\_hand4 sequences. The 3D motion trajectories of the camera in the lab\_hand3 and lab\_hand4 sequences are shown in Fig.~\ref{fig9-exp-3d}, wherein the lost point of ORB-SLAM2 are pointed by the red arrows. In Fig.~\ref{fig9-exp-3d1}, ORB-SLAM2 lost tracking when the camera rotated. In Fig.~\ref{fig9-exp-3d2}, the camera had a violent shake at the place pointed by the arrow, causing ORB-SLAM2 to lose tracking.

In terms of SLAM accuracy, the trajectory error of Multicam-SLAM is lower than that of ORB-SLAM2 except for the lab\_hand3 sequence. The reason why ORB-SLAM2 has a slightly higher trajectory accuracy on the lab\_hand3 sequence is that the untracked camera pose is not included in the calculation. Overall, the handheld experimental results show that Multicam-SLAM proposed in this paper owns higher accuracy and robustness during 3D motions.

\subsubsection{Runtime}

\begin{table}[t]
	\caption{Runtime Evaluation of Multicam-SLAM}
	\label{table-04-runtime}
	\begin{center}
		\begin{threeparttable}
        \setlength{\tabcolsep}{0.6mm}
		\begin{tabular}{c c c c c}
            \toprule
		    \multirow{2}{*}{Task} 	 & \multicolumn{2}{c}{ORB.S2~\cite{mur2017orb}}		& \multicolumn{2}{c}{Multicam.S} \\

            \cmidrule(lr){2-3}\cmidrule(lr){4-5}
            &Median~[ms] &Mean~[ms] &Median~[ms] &Mean~[ms]\\
			\midrule
			Feature Extraction 	    &12.54	&12.78	&18.73	&19.01 \\
			Pose Estimation 		    &3.54	&3.76	&4.39	&5.12 \\
			Local Map Tracking		&12.60	&13.99	&19.24	&20.72 \\
			Total Tracking		&28.68	&30.53	&42.36	&44.85 \\
            \midrule

			Landmark Creation					&71.25	&80.76	&101.87	&107.56 \\
			Local Map BA		    &275.13	&330.66	&337.12	&410.37	\\
			Total Mapping		    &375.30	&450.31	&470.04	&551.17 \\
            \midrule

			Relocalization		&7.30	&7.51	&8.01	&8.72	\\
			Pose-only BA		    &1.37	&1.45	&1.61	&1.73 \\
            \bottomrule
		\end{tabular}
        \begin{tablenotes}
        \item[*] Runtime results are obtained from the handheld dataset experiments.
        \end{tablenotes}
    \end{threeparttable}
	\end{center}
\end{table}

We evaluate the runtime of each thread in the proposed multi-camera visual SLAM method and ORB-SLAM2. Table~\ref{table-04-runtime} shows the computation time of each function in the pose tracking thread and the local mapping thread, in milliseconds. The results show that in the tracking thread, computation is mainly used for feature extraction and tracking of the local map, while in the mapping thread, more time is spent on local BA optimization. Comparing the runtime in the tracking thread and the mapping thread, it can be seen that the time consumption of the mapping and tracking in Multicam-SLAM is higher than that of ORB-SLAM2. This is because multiple RGB-D cameras are used, and the SLAM system needs to process more data. The tracking frame rate of ORB-SLAM2 is about 30 frames per second, while Multicam-SLAM can maintain a tracking speed of 20 frames per second. In the mapping thread, the time consumed by Multicam-SLAM is about 1.24 times that of ORB-SLAM2. Since the mapping thread does not need to run in real time, it can meet the mapping requirements of visual SLAM.

\section{Conclusion} \label{Conclusion}

This paper presented a novel multi RGB-D camera SLAM system, Multicam-SLAM, to improve the accuracy and robustness of visual SLAM. 
The multicam entity is established with multiple cameras with non-overlapping fields of views, providing more comprehensive spatial information. The relative poses between cameras contained in the entity are calibrated on-the-fly with a novel formulation of pose graph optimization regarding the non-overlapping setting. The experimental results under robotic and handheld settings demonstrate the superior accuracy and robustness of the proposed Multicam-SLAM. Future work will focus on further improving the efficiency of the system and exploring its application in more complex environments.

\bibliographystyle{IEEEtran}
\bibliography{IEEEabrv,references}

\begin{IEEEbiography}[{\includegraphics[width=1in,height=1.25in,clip,keepaspectratio]{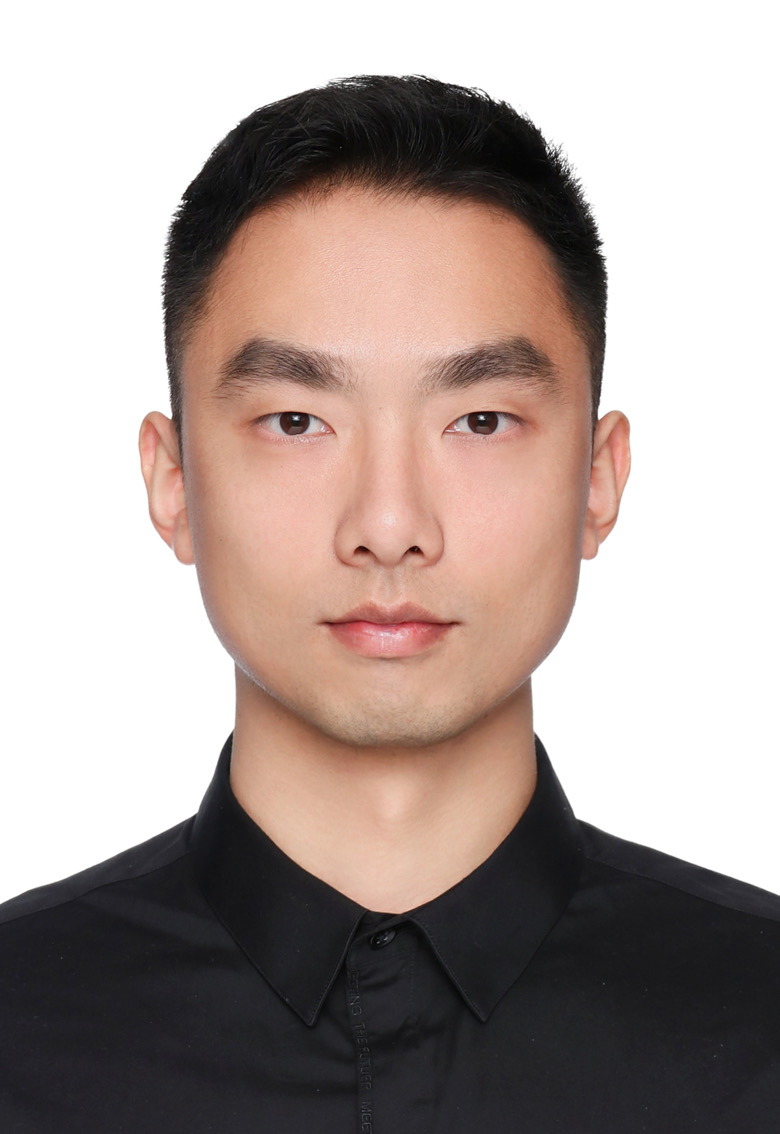}}]{Shenghao Li} received the B.S degree in Mechanical Design manufacture and Automation from East China University of Science and Technology in 2017 and the M.S. degree in Mechanical Engineering in 2020 from East China University of Science and Technology, Shanghai. He is studying for the Ph.D. degree in Control Science and Engineering from Shanghai Jiao Tong University, Shanghai, China. His current research interests include local feature learning, visual localization, and visual-inertial slam.
\end{IEEEbiography}
\begin{IEEEbiography}[{\includegraphics[width=1in,height=1.25in,clip,keepaspectratio]{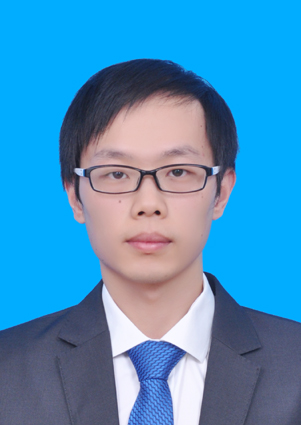}}]{Pang Luchao} received his Bachelor's degree from East China University of Science and Technology in 2015, and his Master's degree from the same university in 2020. During his Master's studies, his research focus was on multi-camera SLAM. Since 2020, he has been working as a software engineer at Arm Technology (China) Co., Ltd. 
\end{IEEEbiography}
\begin{IEEEbiography}[{\includegraphics[width=1in,height=1.25in,clip,keepaspectratio]{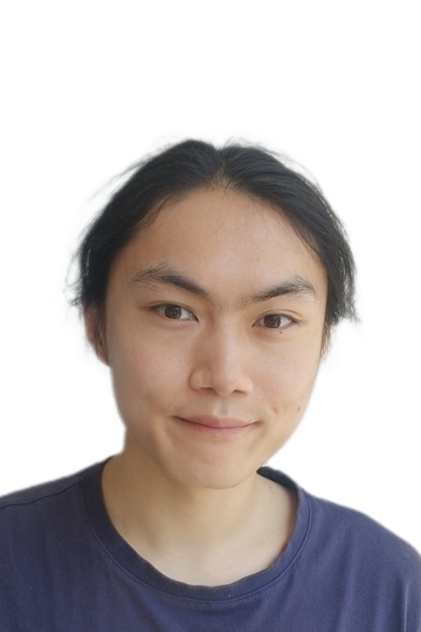}}]{Xianglong Hu} received the B.S degree in Physics from Fudan University in 2017 and the M.S. degree in Computer Science in 2020 from New York Univerisity in New York City. He is currently working in the SaaS tech landscape. His research interests focus on large language model and generative AI. 
\end{IEEEbiography}

\end{document}